\documentclass[letterpaper, 10 pt, conference]{ieeeconf}  

\IEEEoverridecommandlockouts                              
\usepackage{cite}
\usepackage{multirow}
\usepackage{amsmath,amssymb,amsfonts}
\usepackage{algorithmic}
\usepackage{graphicx}
\usepackage{textcomp}
\usepackage{xcolor}
\usepackage{subcaption}

\title{\LARGE \bf
Hybrid Rigid-Soft Robotic Gripper with Shape Adaptation, Uniform Force Distribution, and Self-Locking Capabilities
}
\author{
Xi Chen$^{1,2,\dagger}$, 
Yun Wang$^{1,3,\dagger}$, 
Lichao Yang$^{1,3}$, 
Haitao Li$^{3}$, 
Ya Xiong$^{1,*}$%
\thanks{This work was supported by the Beijing Academy of Agricultural Artificial Intelligence and Robotics - Key Technology Research of Strawberry Harvesting Robot Incorporating Visual-Force Perception and Humanoid Cooperative Operation, the BAAFS Innovation Ability Project (KJCX20240321), the Haidian District Bureau of Agriculture and Rural Affairs, the BAAFS Foundation for Excellent Young Scientists (Grant No. YKPY2025007), and the NSFC Excellent Young Scientists Fund (overseas). (*\textit{Corresponding Author: Ya Xiong, \tt\small yaxiong@nercita.org.cn})}
\thanks{$^\dagger$These authors contributed equally to this work.}
\thanks{$^1$Intelligent Equipment Research Center, Beijing Academy of Agriculture and Forestry Sciences, Beijing, China.}
\thanks{$^2$College of Engineering, China Agricultural University, Beijing, China.}
\thanks{$^3$College of Engineering, Shanxi Agricultural University, Jinzhong, China.}
}

\begin{document}

\maketitle
\thispagestyle{empty}
\pagestyle{empty}

\begin{abstract}

Conventional robotic grippers face a significant challenge in agricultural automation: the trade-off between compliant, adaptive grasping, pressure balancing among all joints, and high load capacity, often at the cost of high energy consumption. This paper presents a novel hybrid rigid-soft gripper that integrated low-cost, membrane-based pneumatic actuators with 3D-printed dual ratchet-pawl mechanisms to simultaneously achieve shape adaptation, uniform force distribution, and energy-free self-locking. The dual-ratchet structure assembled in an offset configuration significantly increased the angular resolution of the joint locking mechanism. Key experimental results demonstrated the gripper's superior performance: a remarkable maximum load capacity of 4200 g, far exceeding that of conventional soft grippers (45–210 g); more uniform force distribution across object sizes (1.75–35.29\% difference ratio) compared to a rigid gripper (56.77–66.44\%), with peak contact forces remaining below surface damage thresholds; and a 50.05\% reduction in total energy consumption to 42.6 J per grasp cycle, achieved by eliminating the need for continuous pneumatic pressure through the self-locking mechanism, compared to 85.28 J for a conventional soft gripper.
The combination of additive manufacturing for ratchets and commercially available materials for pneumatic chambers ensured a low-cost and easily fabricated design. These findings validated that the proposed gripper successfully bridged the gap between soft compliance and rigid reliability, offering a robust and efficient solution for scalable agricultural harvesting and manipulation tasks.
\end{abstract}

\section{INTRODUCTION}

Robotic gripper is a crucial component of robotic systems that plays an increasingly significant role in various fields such as automated manufacturing, healthcare, services, and precision operations\cite{c1}. In the field of agricultural automation, grippers can perform complex tasks such as fruit harvesting and sorting. Enhancing the flexibility, adaptability, and efficiency of robotic grippers has attracted considerable attention in the field of agricultural robotics\cite{c2}.

\begin{figure}[htbp]
    \centering
    \includegraphics[width=0.9\linewidth]{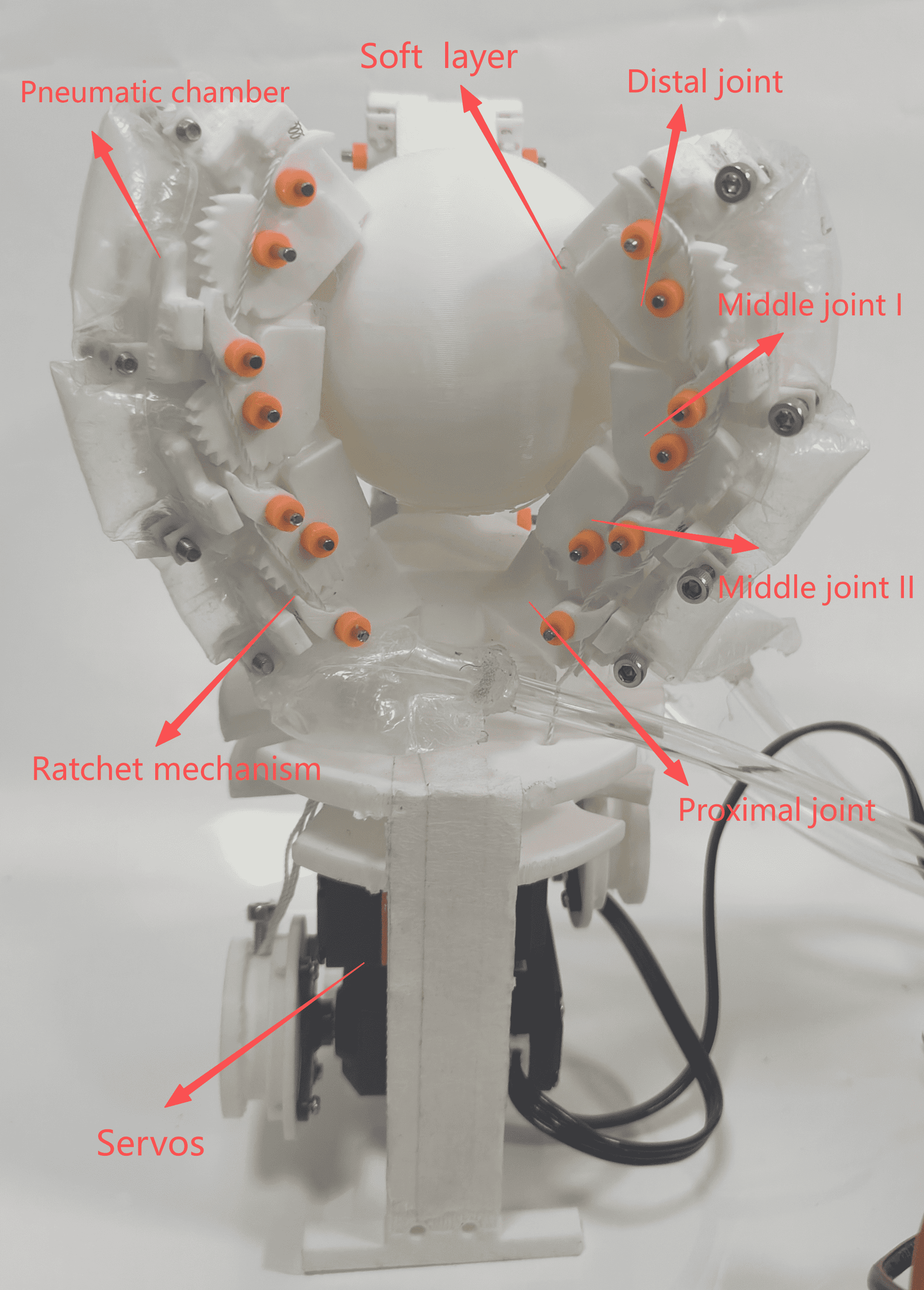}
    \caption{The proposed hybrid rigid–soft robotic gripper.}
    \label{fig:gripper}
\end{figure}

Currently, traditional robotic grippers are predominantly designed with rigid structures, which excel in precision and force output \cite{c3}. However, rigid robotic grippers struggle to grasp irregular, fragile, or geometrically complex objects, and their grasping capability is often insufficient when dealing with low-load or non-standard objects \cite{c4}. This limitation becomes particularly critical in unstructured agricultural scenarios, such as fruit and vegetable harvesting, where target objects are delicate, shape-varying, and often occluded by leaves and branches. Additionally, the lack of adaptability in rigid structures leads to limited flexibility when responding to dynamic environmental changes, ultimately affecting the efficiency and stability of task execution \cite{c4}. Therefore, particularly for agricultural picking tasks involving damage-sensitive crops, enhancing the flexibility and adaptability of robotic grippers has emerged as a pivotal research focus \cite{c5}.

Compared to rigid robotic grippers, the design of soft robotic grippers offers clearer advantages in terms of adaptability and flexibility \cite{c6}, \cite{c9}. Owing to their inherent mechanical compliance, soft grippers can conform to object shapes and surface characteristics without complex control inputs, enabling adaptive grasping while reducing system complexity and improving safety \cite{c7}. This compliance allows the gripper to passively deform during grasping, thereby achieving gentler contact and effectively reducing impact forces when handling fragile or irregularly shaped objects \cite{c8}, \cite{c10}. Such characteristics make soft grippers promising candidates for agricultural picking tasks that require safe interaction with damage-sensitive crops.Despite these advantages, soft robotic grippers face challenges in high-load tasks and precise control \cite{c11}. Due to insufficient structural rigidity, they may struggle to generate adequate force during heavy-duty operations and can exhibit instability and reduced accuracy, particularly when grasping hard or heavy objects \cite{c12}, \cite{c13}. These limitations may also affect operational reliability in agricultural environments, where variability in crop size, attachment strength, and environmental conditions demands both compliance and sufficient force output.


In recent years, hybrid rigid–soft grippers have emerged to combine the advantages of both approaches. However, existing hybrid designs still face several limitations. For instance, the gripper developed by Zhu et al. demonstrated high payload capacity and dexterous in-hand manipulation, but exhibited limited lateral compliance in certain grasping directions\cite{c14}. The origami-based hybrid gripper proposed by Zhang et al. achieved variable stiffness, yet its force control precision and structural robustness were insufficient when handling heavy or dynamically varying loads\cite{c15}. The RISOs gripper enhanced versatility and object diversity but its performance under prolonged loading and energy efficiency remains insufficiently validated\cite{c16}. 



To address these challenges, this study proposed a hybrid rigid–soft gripper combining a dual ratchet–pawl self-locking mechanism with segmented pneumatic chambers. The ratchet–pawl structure enabled passive joint locking and high load-bearing capacity without continuous actuation, while the dual configuration improved resolution and robustness. Pneumatic chambers were fabricated using low-cost nylon–polyethylene composite membranes arranged in segments, providing compliant shape conformation with simplified assembly. The main contributions are as follows:
\begin{enumerate}

\item A hybrid rigid–compliant gripper architecture that integrated a ratchet–pawl locking mechanism with segmented pneumatic chambers to achieve both structural stability and adaptive grasping.

\item A low-cost membrane-based pneumatic chamber design using nylon–polyethylene composite materials, enabling simplified fabrication and maintenance.

\item A dual ratchet–pawl self-locking mechanism that improved joint locking resolution, increased load capacity, and reduced energy consumption.

\end{enumerate}

Overall, the proposed gripper achieved a balance between compliance and load-bearing capability, enabling practical deployment in agricultural harvesting and similar large-scale manipulation scenarios requiring robustness and cost-effectiveness.

\section{Design}
The proposed robotic gripper comprised three fingers, as shown in Fig.~\ref{fig:gripper}. Each finger consisted of four joints, incorporating four rigid links, three compliant pneumatic actuators, and three dual ratchet-based self-locking mechanisms. Two servo actuators were used to disengage the ratchet-pawl mechanisms. Each joint module had a length of 23 mm, a width of 25 mm, and a height of 23 mm, resulting in a compact structural configuration. The design of each module was coordinated with all other components to guarantee efficient and stable system operation during task execution.

\subsection{Finger Design}

The proposed finger joint mechanism integrated compliant pneumatic actuation units with rigid structural components to achieve grasping stability while providing a certain degree of adaptability. The pneumatic chambers were fabricated from thin plastic membranes, which undergo significant deformation during pressurization to drive joint bending and conform to the geometry of target objects. In contrast, the rigid components ensured structural support and positional accuracy under high-load conditions.

During grasping tasks, the four joints of each finger bent under pneumatic actuation, enabling the fingers to conform to the object shape and achieve adaptive grasping. Since each joint exerted a relatively low force during actuation, the design effectively mitigated the local stress concentration and reduced the risk of surface damage, thus enhancing the suitability for handling delicate targets such as fruits\cite{c17}.

The joints were connected in series, each joint providing an independent rotational degree of freedom, which ensured high compliance and general adaptability of the finger. In addition, the mechanical design of the joint axes reduced friction and wear, contributing to the long-term structural reliability and durability of the gripper.


\subsection{The Dual-Ratchet Pawl Mechanism Design}

\begin{figure}[htbp]
    \centering
    \includegraphics[width=1\linewidth]{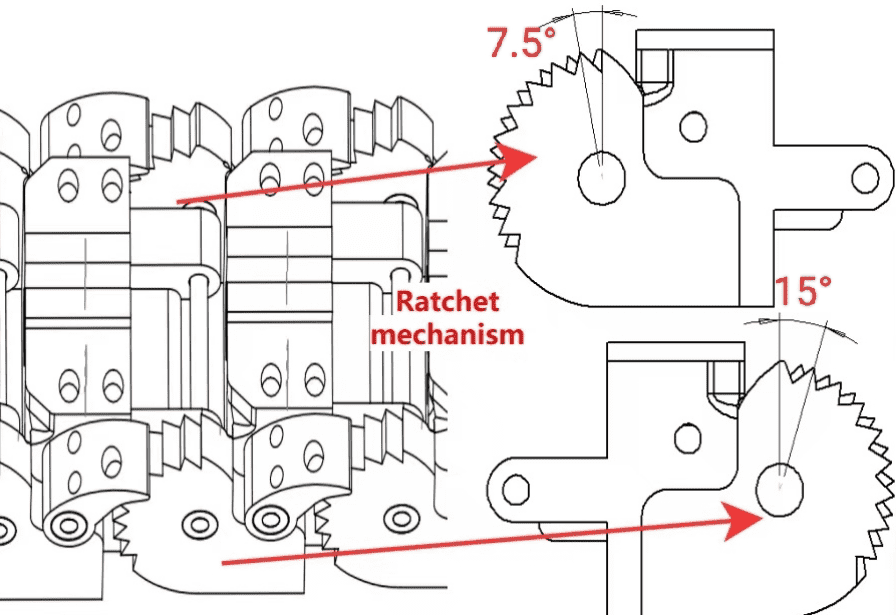}
    \caption{Ratchet mechanisms mounted on both sides of each joint.}
    \label{fig:2}
\end{figure}

To reduce energy consumption and increase the grasping load capacity of the gripper, we designed a dual-ratchet pawl self-locking mechanism for each joint. A self-locking mechanism is a structure that restricts relative motion once engagement occurs, thereby maintaining the system's state without requiring continuous external actuation. Once a joint reaches its target bending position, these mechanisms enable reliable self-locking, allowing the gripper to maintain grasp stability without continuous pneumatic power. Unlike conventional joints that rely on sustained actuation, the ratchet–pawl mechanisms preserved the joint position through mechanical engagement, effectively preventing loosening or slippage caused by pneumatic fluctuations or external 
disturbances\cite{c18}. Conventional single-ratchet mechanisms suffer from limited rotation angle resolution, making it difficult to achieve precise locking after the robotic gripper adapts to an object's shape. To address this limitation, the dual-ratchet structure was introduced, increasing the resolution of the ratchet rotation angle, as illustrated in Fig.~\ref{fig:2}. In this design, both ratchets within a single system featured identical tooth count and pitch, but they were assembled in an offset configuration. For example, the single-ratchet rotation resolution was 15°, while the dual-ratchet combination improved it to 7.5°, effectively doubling the resolution. This structural enhancement significantly increased the precision and stability of the locking mechanism. The pawls were spring-loaded to ensure consistent contact with the ratchet teeth during joint motion, thereby achieving rapid and robust locking under diverse grasping conditions.

\begin{figure}[htbp]
    \centering
    \includegraphics[width=1\linewidth]{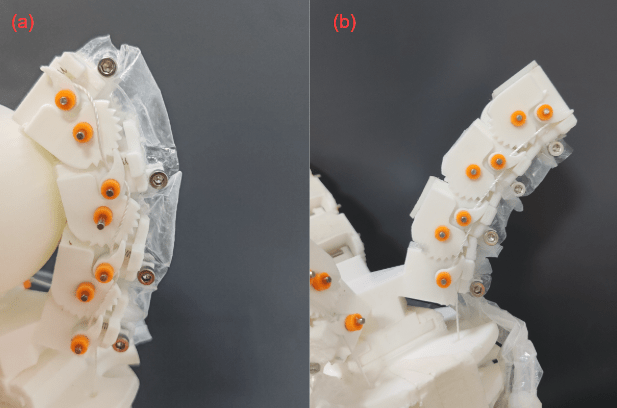}
    \caption{(a) The pawl engaged with the ratchet teeth during grasping, maintaining the joint in a locked state; (b) the pawl disengaged from the ratchet teeth during resetting, allowing the joint to enter an unlocked state and gradually return under actuation.}
    \label{fig:3}
\end{figure}

The reset process is illustrated in Fig.~\ref{fig:3}. During grasping, the pawl engages with the ratchet teeth to maintain joint locking. During resetting, a servo motor tightens a cable to disengage the pawl from the ratchet and subsequently drives the joint back to its initial position. The cable must exhibit high flexibility to accommodate joint motion and sufficient tensile strength and toughness to prevent rupture under repeated loading. In this design, a commercially available 12-strand cable was employed to satisfy these requirements and ensure reliable long-term operation.

In addition, to reduce manufacturing cost and simplify assembly, the ratchet components were fabricated using additive manufacturing. This approach provided sufficient structural strength for the intended application while facilitating rapid prototyping and iterative design optimization.

\subsection{Low-Cost Pneumatic Chambers}

The pneumatic actuation system was constructed using low-cost, commercially available check-valve membranes made of a nylon–polyethylene composite. This design offered several advantages, including low cost, ease of fabrication, and convenient assembly or replacement, while also reducing reliance on high-precision manufacturing processes. Compared with conventional chambers that required dedicated molds or precise machining, the membrane-based design significantly lowered sensitivity to manufacturing accuracy. The material was not only inexpensive but also simplified assembly and allowed for quick replacement, enhancing maintainability.
Moreover, the segmented configuration provided sufficient compliance to conform to diverse object geometries and improving adaptability for handling fragile agricultural products.

\begin{figure}[htbp]
    \centering
    \includegraphics[width=1\linewidth]{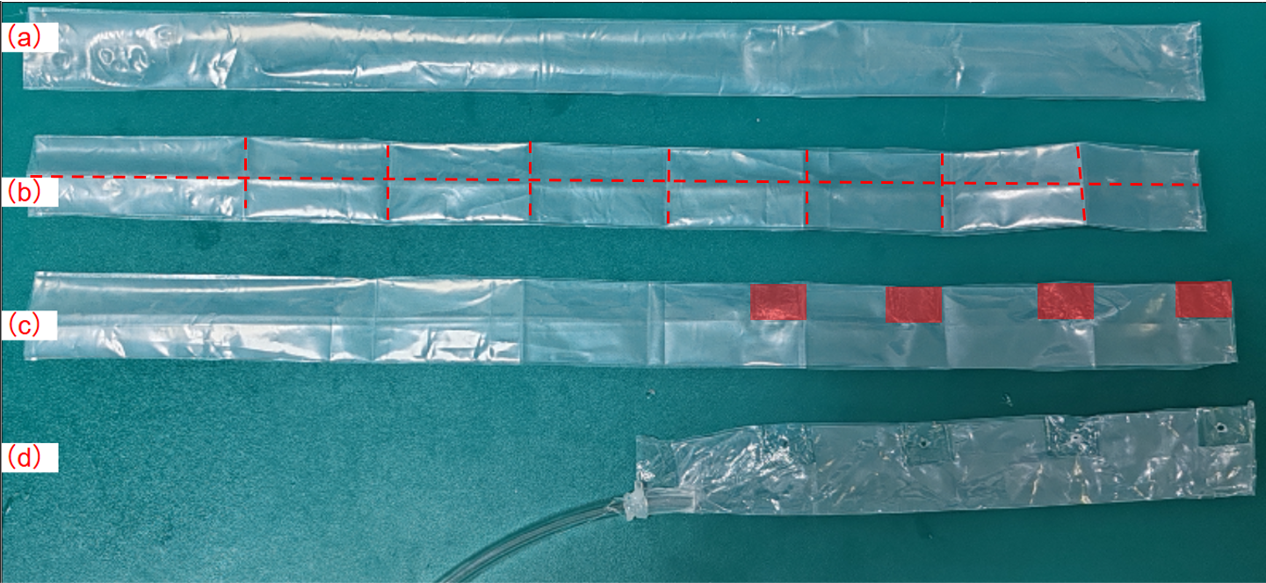}
    \caption{Fabrication process of the air chamber: (a) purchased strip-shaped check-valve membrane; (b) folding and marking according to the designed length, with each chamber segment set to 45 mm; (c) partial sealing along the marks using a heat sealer, leaving regions for perforation; (d) punching at the sealed sections, trimming the excess membrane, and sealing the tail end to a PU air tube with hot-melt adhesive.}
    \label{fig:4}
\end{figure}

The fabrication process is illustrated in Fig.~\ref{fig:4}. First, strip-shaped check-valve membranes were purchased, as shown in Fig.~\ref{fig:4}(a). The required lengths for each finger segment of the gripper were then measured, and folds were made as markers. In this design, each chamber segment was set to a length of 45 mm (indicated by the dashed line in Fig.~\ref{fig:4}(b)). Next, partial sealing was performed along the marked positions using a heat sealer (highlighted in red in Fig.~\ref{fig:4}(c)), leaving regions for perforation to facilitate subsequent fixation onto the gripper structure. Finally, a hole was punched at the center of each sealed section, excess material was trimmed, and the tail end was sealed and fixed to a TPU air tube with hot-melt adhesive, thereby completing the fabrication of the air chamber (as shown in Fig.~\ref{fig:4}(d)).

\section{Experiments and Results}
\subsection{Experimental Setup}\label{GFT}

\begin{figure}[htbp]
    \centering
    \includegraphics[width=1\linewidth]{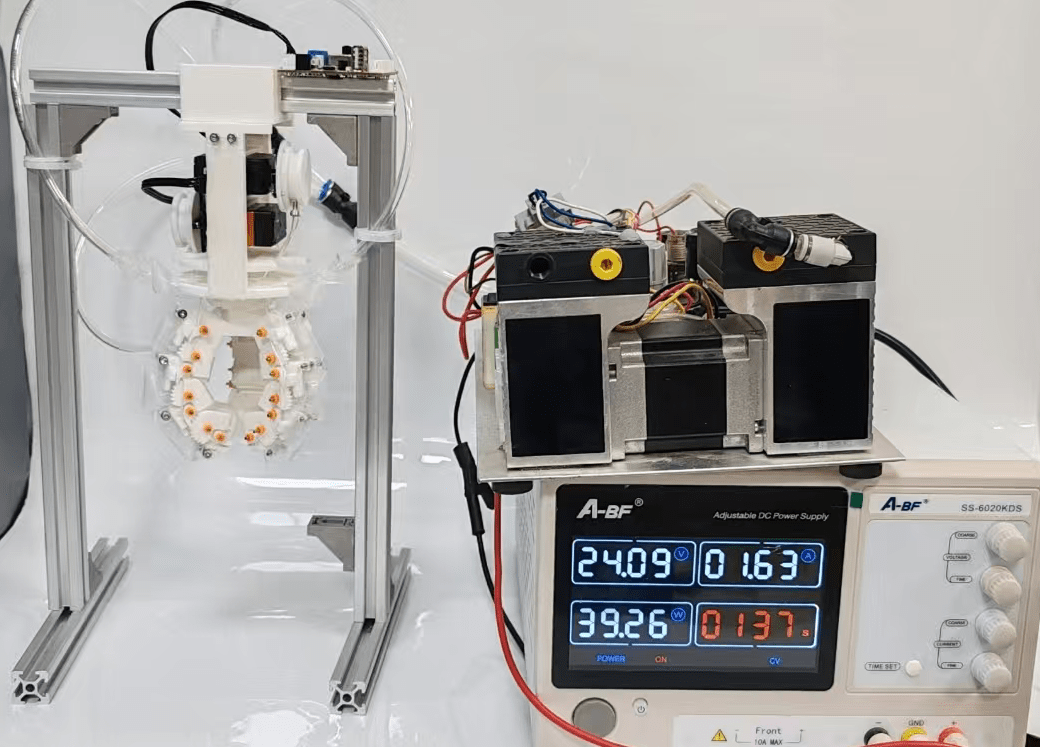}
    \caption{Experimental setup for demonstrating the proposed gripper.}
    \label{fig:6}
\end{figure}

The experimental setup is illustrated in Fig.~\ref{fig:6}. Our gripper was mounted on a test bench constructed from aluminum profiles. The pneumatic source was supplied by an air pump (AD30KEEDB24PHV, Shanghai, China), and the power source was provided and recorded by a digital DC power supply (A-BF SS3020KDS, Mainland China). The contact surfaces between the joints and the target objects were covered with a silicone layer, in which resistive thin-film pressure sensors with a diameter of 8 mm (RP-C18.3-LT, K-CUT, Mainland China) were embedded. These sensors were connected to an Arduino UNO control board, and the experimental data were collected and recorded via an upper computer. The spherical models used in the experiments were fabricated using FDM 3D printing with diameters of 50 mm, 60 mm, and 70 mm. PLA basic filament was used, and the infill was set to 15\% to ensure lightweight structures.

\subsection{Grasping Force Test}

Experiment I evaluated the joint contact pressures during grasping to assess compliance and load distribution when handling fragile objects. Resistive thin-film pressure sensors (diameter: 8 mm) were embedded at the contact interfaces between the joints and the object surface. Spherical models with diameters of 50 mm, 60 mm, and 70 mm were fabricated via FDM 3D printing using PLA (15\% infill).

\begin{figure}[htbp]
    \centering
    \includegraphics[width=1\linewidth]{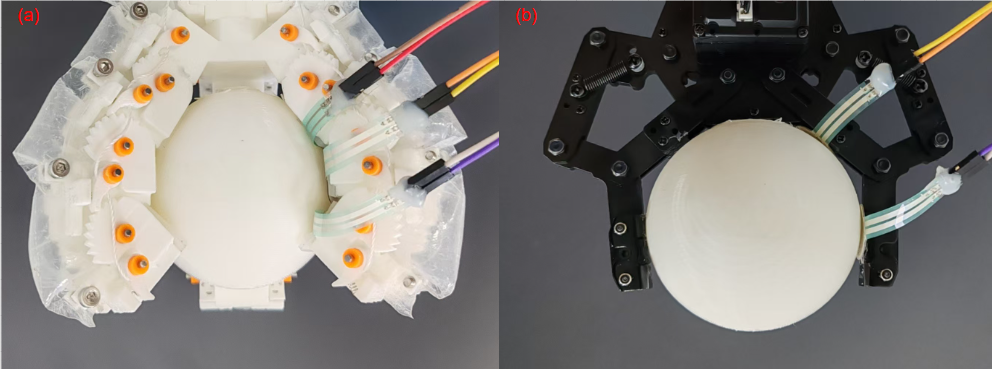}
    \caption{Contact pressure measurements of different joints using thin-film sensors.
(a) Proposed gripper; (b) rigid gripper.}
    \label{fig:77}
\end{figure}

At an input pressure of 5 kPa, the proposed gripper grasped the spherical models, and joint contact pressures were recorded. For comparison purposes, a rigid parallel-link gripper was tested under stable grasping conditions of the same objects (Fig.~\ref{fig:77}).

Each condition was repeated 11 times under identical settings, and the mean and sample standard deviation of the joint contact pressures were calculated. Within the pressure range required for stable grasping, the overall grasping force of the proposed gripper showed limited sensitivity to input pressure variations; therefore, tests under different pneumatic pressures were not conducted.

To quantify the uniformity of load distribution between joints, a Difference Ratio was defined as

\[
\text{Difference Ratio} = \frac{|F_1 - F_2|}{F_1} \times 100\%,
\]

where \(F_1\) denotes the larger mean pressure and \(F_2\) the smaller one. For each object diameter, the ratio was computed using the distal and Middle II joint mean pressures.

\begin{table}[htbp]
    \centering
    \caption{Joint contact pressure statistics of our gripper and a rigid parallel-linkage gripper when grasping spherical models of different diameters.}
    \label{tab:pressure_stats}
    \begin{tabular}{p{1.0cm} p{0.8cm} p{1.2cm} p{1.0cm} p{1.0cm} p{1.0cm}}
        \hline
\textbf{Gripper} & \textbf{Diameter} & \textbf{Joint} & \textbf{Mean} & \textbf{Std} & \textbf{Diff.} \\
                 & \textbf{(mm)}     &                & \textbf{(N)}  & \textbf{(N)} & \textbf{Ratio} \\
        \hline
        
        \multirow{9}{*}{\shortstack{Our\\gripper}}
        & \multirow{3}{*}{50}
        & Distal    & 0.33 & 5.94 & \multirow{3}{*}{35.29\%} \\
        &  & Middle I    & 0.00 & 0.00 &  \\
        &  & Middle II & 0.51 & 7.67 &  \\
        
        & \multirow{3}{*}{60}
        & Distal   & 0.56 & 7.98 & \multirow{3}{*}{1.75\%} \\
        &  & Middle I  & 0.00 & 0.00 &  \\
        &  & Middle II  & 0.57 & 8.30 &  \\
        
        & \multirow{3}{*}{70}
        & Distal   & 0.50 & 9.57 & \multirow{3}{*}{16.00\%} \\
        &  & Middle I   & 0.00 & 0.00 &  \\
        &  & Middle II  & 0.42 & 5.85 &  \\
        
        \hline
        
        \multirow{9}{*}{\shortstack{Parallel\\rigid\\gripper}}
        & \multirow{3}{*}{50}
        & Distal    & 1.55 & 26.42 & \multirow{3}{*}{56.77\%} \\
        &  & Middle I  & 0.00 & 0.00 &  \\
        &  & Middle II  & 0.67 & 11.19 &  \\
        
        & \multirow{3}{*}{60}
        & Distal    & 1.49 & 12.49 & \multirow{3}{*}{66.44\%} \\
        &  & Middle I   & 0.00 & 0.00 &  \\
        &  & Middle II  & 0.50 & 7.77 &  \\
        
        & \multirow{3}{*}{70}
        & Distal    & 2.60 & 36.94 & \multirow{3}{*}{62.69\%} \\
        &  & Middle I   & 0.00 & 0.00 &  \\
        &  & Middle II  & 0.97 & 25.84 &  \\
        
        \hline
    \end{tabular}
\end{table}

The statistical results are summarized in Table~\ref{tab:pressure_stats}. The reported standard deviations correspond to the sample standard deviation of the repeated trials.

Across all object diameters, the proposed gripper exhibited lower contact pressures and substantially smaller distal–Middle II Difference Ratios than the rigid gripper, indicating improved load sharing.

For the 50 mm sphere, the proposed gripper distributed contact between the distal (0.33 N) and Middle joint II (0.51 N), resulting in a Difference Ratio of 35.29\%. In contrast, the rigid gripper showed pronounced distal concentration (1.55 N vs. 0.67 N), yielding a higher Difference Ratio of 56.77\% and greater variability (Std = 26.42 N).

For the 60 mm sphere, the proposed gripper achieved nearly uniform load sharing (0.56 N vs. 0.57 N; Difference Ratio = 1.75\%), whereas the rigid configuration remained strongly imbalanced (1.49 N vs. 0.50 N; Difference Ratio = 66.44\%).

At 70 mm, the proposed design maintained moderate and relatively balanced engagement (0.50 N vs. 0.42 N; Difference Ratio = 16.00\%), while the rigid gripper exhibited significant distal dominance (2.60 N vs. 0.97 N; Difference Ratio = 62.69\%) and the largest observed standard deviation (36.94 N).

Overall, the hybrid rigid–soft mechanism promoted more uniform joint participation and reduced pressure concentration compared with the rigid configuration. The consistently lower Difference Ratios and reduced dispersion across object sizes demonstrated improved load-sharing stability. The relatively larger Difference Ratios observed for the rigid gripper can be attributed to its two-finger grasping configuration, which provides less geometric constraint when grasping spherical objects. In contrast, the three-finger arrangement of the proposed gripper forms a more stable grasping structure, improving contact distribution and grasp stability.

\subsection{Grasping Force Tests under External Loads}
Experiment II evaluated the pressure responses of different joints when the gripper, after enveloping the target object, was subjected to a unidirectional external force (e.g., fruit–peduncle separation during harvesting). The procedure was as follows. First, the pneumatic source supplied an input pressure of 5 kPa to achieve firm grasping of the spherical model. The source was then turned off, and the gripper maintained grasping through the ratchet-based self-locking mechanism. Subsequently, a longitudinal external load was gradually applied to the model, and the joint pressures were recorded.

\begin{figure}[htbp]
    \centering
    \includegraphics[width=1\linewidth]{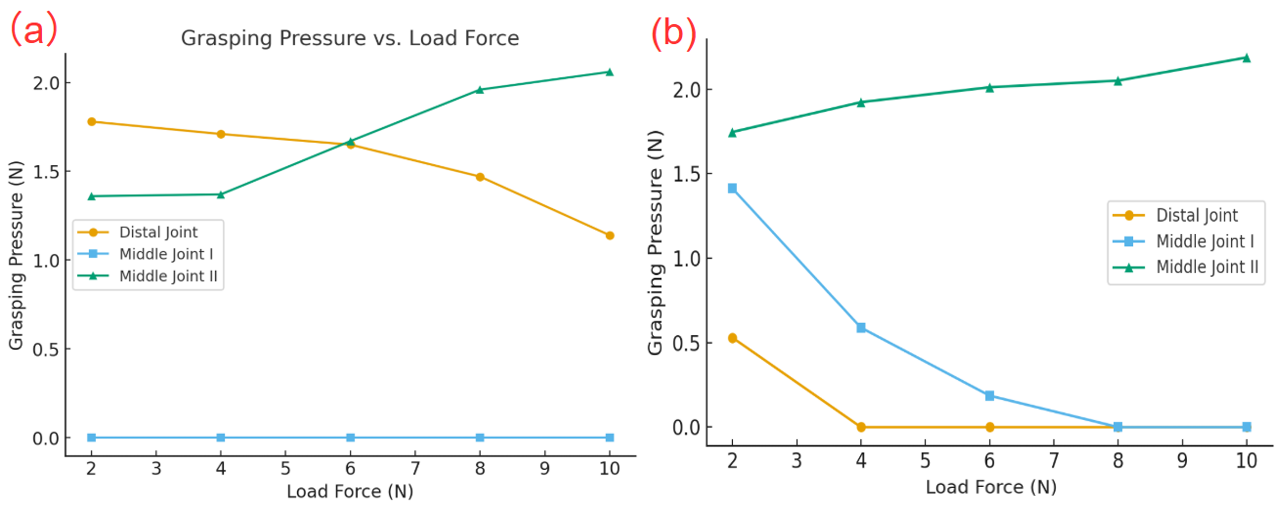}
    \caption{Relationship between joint grasping pressures and applied external loads during the self-locking stage after pneumatic actuation was cut off: (a) Grasping a 50 mm spherical model; (b) Grasping a 60 mm spherical model. Each plot shows three curves corresponding to the distal joint, middle joint I, and middle joint II.}
    \label{fig:8}
\end{figure}

The direction of the applied external load was perpendicular to the gripper base and directed outward from the base, simulating a pulling force along the longitudinal axis of the grasped object. The external load was monitored using a force gauge (SMF-10, Mainland China), starting from 2 N and increasing in increments of 2 N up to 10 N. The relationship curves between joint pressures and external loads were obtained when grasping 50 mm and 60 mm diameter spherical models, as shown in Fig.~\ref{fig:8}.


The results showed that immediately after grasping, the 50 mm sphere primarily experienced pressures from the distal joint and middle joint II (closer to the finger root), while the 60 mm sphere mainly experienced pressures from the distal joint and middle joint I (closer to the finger tip). As the external pulling force increased, the pressure at the distal joint rose, whereas the pressures at the other joints gradually decreased. The results demonstrated that, for objects within a certain size range, the joints of our gripper maintained a relatively balanced force distribution under applied external loads.

\subsection{Load Capacity Test}\label{LCT}
\begin{figure}[htbp]
    \centering
    \includegraphics[width=1\linewidth]{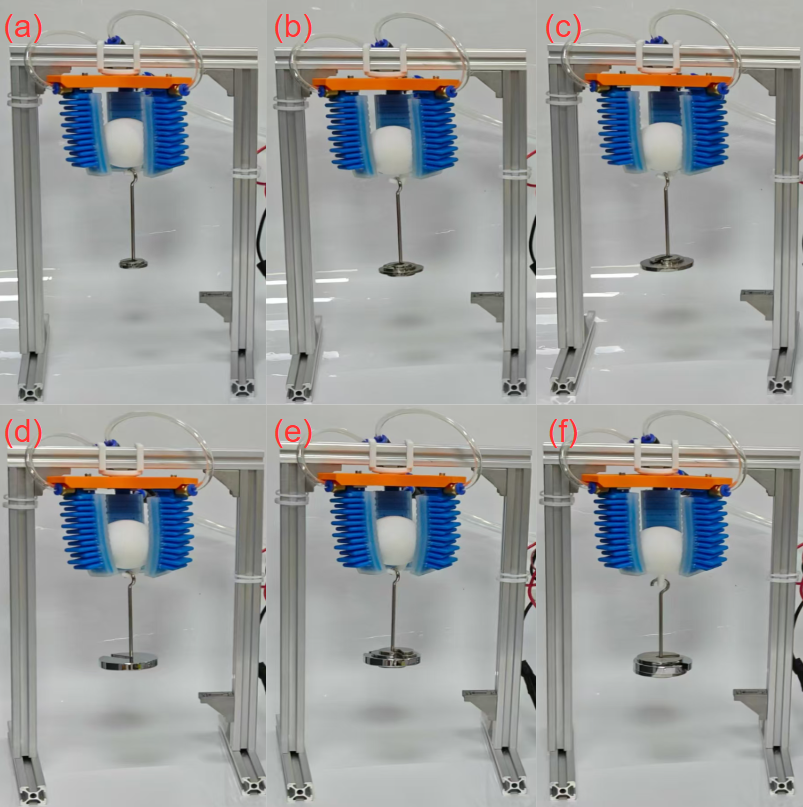}
    \caption{Maximum load capacity of a conventional soft gripper measured under different pneumatic pressures: (a) 5 kPa, (b) 10 kPa, (c) 15 kPa, (d) 20 kPa, (e) 25 kPa, and (f) 30 kPa.}
    \label{fig:9}
\end{figure}

Experiment III evaluated the load-bearing capacity of the proposed gripper in comparison with that of a conventional soft gripper. For the conventional gripper, the researchers tested its maximum load capacity under pneumatic pressures of 5, 10, 15, 20, 25, and 30 kPa by gradually increasing the mass of the weights (see Fig.~\ref{fig:9}). The weights were added in increments of 5 g, and the model ball had a mass of 20.14 g. During the test, the conventional gripper was fixed on the experimental platform and supplied with the same air source as the proposed gripper. It tightly grasped a lightweight 50 mm diameter model ball, while weights were suspended below the ball to measure the load capacity.

Unlike the conventional structure, the load-bearing capacity of the proposed gripper mainly depended on the locking strength of the ratchet self-locking mechanism rather than pneumatic pressure. Therefore, the pneumatic pressure was uniformly set at 5 kPa during the test. After the gripper grasped the model ball, the air supply was turned off, and the gripper relied solely on the self-locking mechanism to maintain the grasp. The results indicated that the proposed gripper exceeded the total mass of the available weight set.

\begin{figure}[htbp]
    \centering
    \includegraphics[width=0.7\linewidth]{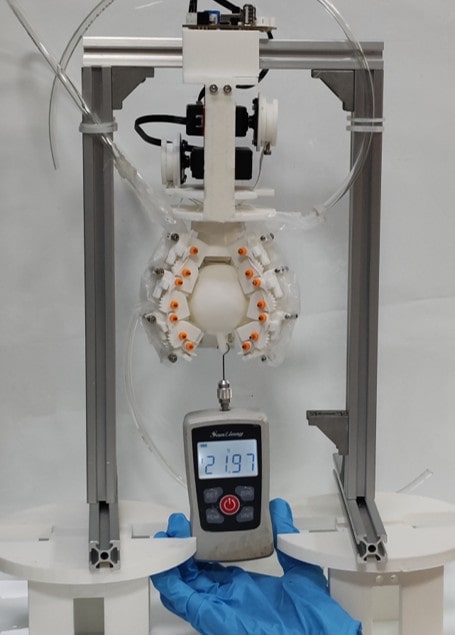}
    \caption{Destructive testing of the proposed gripper to determine its maximum load capacity.}
    \label{fig:111}
\end{figure}

Subsequently, the researchers conducted a destructive experiment using a dynamometer to determine the maximum load capacity and failure mode of the gripper (see Fig.~\ref{fig:111}). The dynamometer was connected to the model ball, and a downward tensile force was gradually applied until the gripper failed. The maximum value recorded by the dynamometer was defined as the load-bearing limit of the proposed gripper.

\begin{table}[htbp]
\centering
\caption{Load capacity of the conventional soft gripper and the proposed gripper.}
\label{tab:capacity}
\renewcommand{\arraystretch}{1} 
\fontsize{6}{12}\selectfont     
\begin{tabular}{lcccccc}
\hline
\textbf{Pressure (kPa)} & 5 & 10 & 15 & 20 & 25 & 30 \\
\hline
\textbf{Conventional soft gripper (g)} & 45.14 & 75.14 & 105.14 & 140.14 & 170.14 & 210.14 \\
\textbf{Proposed gripper (g)} & 4200 & - & - & - & - & - \\
\hline
\end{tabular}
\end{table}

The experimental results were summarized in Table 2. The conventional soft gripper demonstrated maximum load capacities of 45.14 g, 75.14 g, 105.14 g, 140.14 g, 170.14 g, and 210.14 g at pneumatic pressures of 5, 10, 15, 20, 25, and 30 kPa, respectively. In contrast, the proposed gripper demonstrated a maximum load capacity of approximately 4200 g in the destructive test, with failure occurring at the joint connections. These results highlighted that the proposed gripper had a significantly superior load-bearing capability compared with the conventional soft gripper, making it highly valuable for grasping heavy objects or harvesting hard-to-pick fruits.

\subsection{ Energy Consumption Test}\label{ECT}
Experiment IV evaluated the energy consumption of the gripper during operation using a digital DC power supply (A-BF SS3020KDS, China). The operation of the gripper was divided into three stages: 1) pneumatic actuation stage: the joints were pneumatically actuated to grasp the target object, lasting approximately 0.8 s; 2) self-locking holding stage: the air supply was cut off, and the ratchet self-locking mechanism maintained the grasp without additional energy input; 3) reset stage: a servo motor pulled the cord to disengage the ratchet pawl, returning the joints to their initial positions, lasting approximately 2 s.

\begin{figure}[htbp]
    \centering
    \includegraphics[width=1\linewidth]{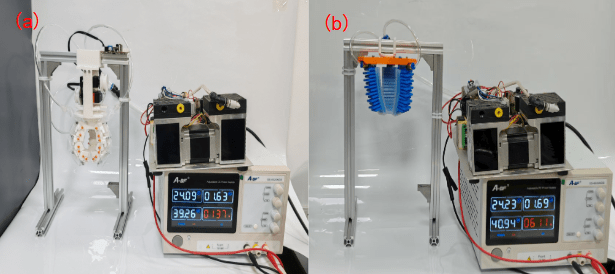}
    \caption{Energy consumption comparison during operation between the proposed gripper and a conventional soft gripper.
(a) Proposed gripper; (b) Conventional soft gripper.}
    \label{fig:10}
\end{figure}

Power consumption during the first and third stages was recorded, and the total energy consumption was calculated. For comparison, the average power consumption of a conventional soft gripper under continuous pneumatic actuation was also measured (Fig.~\ref{fig:10}).

\begin{figure}[htbp]
    \centering
    \includegraphics[width=1\linewidth]{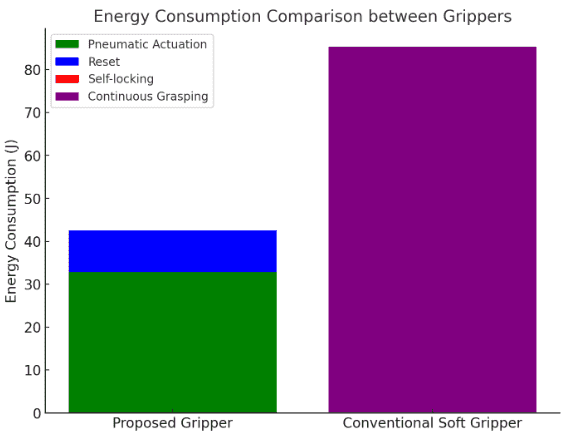}
    \caption{Comparison of energy consumption between the proposed gripper and a conventional soft gripper during operation.}
    \label{fig:11}
\end{figure}

As shown in Fig.~\ref{fig:11}, the proposed gripper consumed an average of 41.0 W during the pneumatic actuation stage (0.8 s) and 4.9 W during the reset stage (2 s). No additional energy was required during the self-locking holding stage, resulting in a total energy consumption of approximately 42.6 W. In contrast, the conventional soft gripper required continuous energy input during grasping, with an average power of 41.0 W and a total energy consumption of 85.28 W over a duration of 2.08 s, nearly twice that of the proposed gripper. These results demonstrated that the proposed gripper demonstrated significant energy efficiency, which was particularly valuable for agricultural applications with limited power supply.

\subsection{Fruit Grasping and Harvesting Tests}
\begin{figure}[htbp]
    \centering
    \includegraphics[width=1\linewidth]{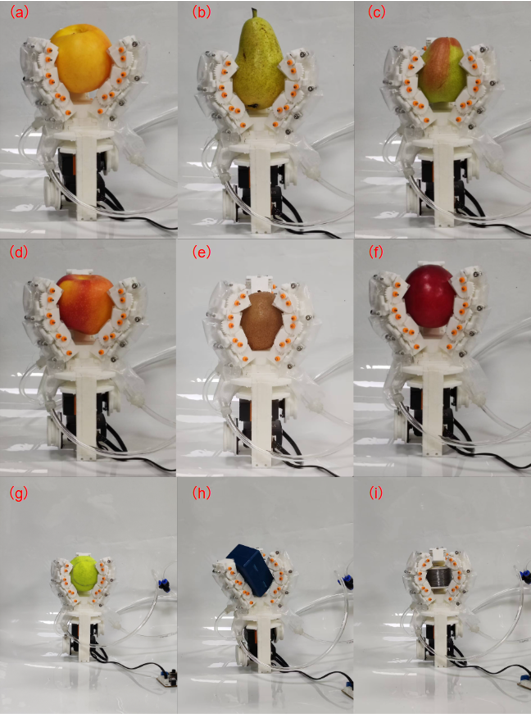}
    \caption{Grasping demonstrations of the proposed gripper on various real objects: (a) yellow peach; (b) pear; (c) July red pear; (d) apple; (e) kiwi; (f) plum; (g) tennis ball; (h) square box; (i) cylindrical coil.}
    \label{fig:12}
\end{figure}

Experiment V evaluated the adaptability and versatility of the proposed gripper through fruit grasping and harvesting tests. The grasping test objects included six types of fruits (yellow peach, pear, July red pear, apple, kiwifruit, and plum) and three daily objects with different shapes (a tennis ball, a square box, and a cylindrical coil), as shown in Fig.~\ref{fig:12}. These objects varied significantly in material, surface characteristics, and geometry, providing a comprehensive evaluation of the gripper’s adaptability in handling both complex agricultural products and general objects.

During the experiments, each type of object was grasped five times. The gripper consistently achieved stable grasps without any additional adjustments, and no noticeable surface damage to the fruits was observed, indicating its compliance and non-destructive characteristics.

For deformable fruits such as yellow peaches, unlike rigid grippers that required precise control of joint positions, the pneumatic actuation allowed the joints to naturally conform to the object surface, thereby reducing contact forces during grasping. In addition, the multi-joint design provided a larger effective contact area, resulting in more uniform pressure distribution and lower peak stress on the fruit surface. These features highlighted the gripper’s capability to interact effectively with deformable objects through compliant pneumatic enveloping combined with distributed contact.

When grasping irregular objects such as the square box and cylindrical coil, stable holding was also achieved in all five trials for each object.

\begin{figure}[htbp]
    \centering
    \includegraphics[width=1\linewidth]{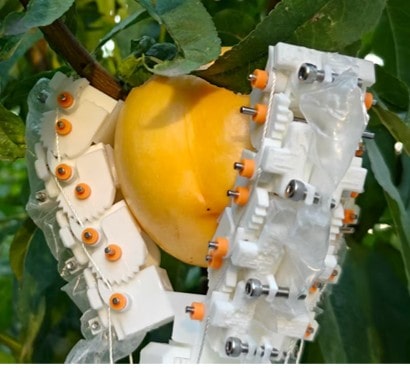}
    \caption{Demonstration of the proposed gripper in harvesting of a yellow peach.}
    \label{fig:13}
\end{figure}

Furthermore, a field harvesting experiment with yellow peaches was conducted once (see Fig.~\ref{fig:13}) to verify the gripper’s performance under real harvesting conditions. The gripper successfully completed the grasping task without fruit dropping or surface damage, demonstrating its effectiveness in field applications.

Overall, the results confirmed that the proposed gripper demonstrated broad adaptability across diverse target objects and possessed the potential for non-destructive harvesting of delicate fruits, providing a solid foundation for future deployment in agricultural automation.

\section{DISCUSSION}
The proposed gripper demonstrated reliable grasping performance through the combination of pneumatic compliance and ratchet-based self-locking. However, several limitations should be acknowledged. Since all joints were actuated simultaneously under pneumatic input, the bending motion occurred concurrently along the finger structure. As a result, the gripper achieved optimal contact when interacting with relatively rounded objects, where uniform enveloping can be naturally established. In contrast, during parallel grasping configurations or when handling objects with concave surface features, the fingers may not achieve tight geometric conformity. This limitation originates from the absence of independent joint control and the inherently coupled actuation strategy.

In addition, the gripper operated effectively within a finite size range of target objects. Objects that were excessively small or excessively large relative to the finger length and base spacing may not be enveloped securely. Nevertheless, the finger structure adopted a modular design. The grasping range can be adjusted by modifying the base dimensions of the gripper and by altering the number of joints per finger. This modularity provided structural flexibility for adapting the design to different application scenarios without fundamentally changing the core mechanism.

\section{CONCLUSIONS}
This study presented the design, fabrication, and experimental validation of a hybrid rigid–soft robotic gripper that addresses the conflicting requirements of adaptive compliance, high load capacity, and energy efficiency in agricultural applications. The proposed system integrated three key components within a unified architecture: (1) segmented membrane-based pneumatic chambers for low-cost and conformable grasping, (2) a spatially adaptive multi-joint finger structure for distributed contact and stable force transmission, and (3) a dual ratchet–pawl self-locking mechanism for high-resolution, passive joint locking without continuous energy input.
Experimental results validated the effectiveness of the proposed hybrid design. The gripper achieved reliable shape-adaptive grasping across a range of objects, including delicate fruits and daily items, while maintaining maximum contact forces below the damage threshold of fragile produce such as peaches (1.8 N). The dual ratchet–pawl mechanism enabled a load capacity of approximately 42 N—significantly higher than that of conventional soft grippers—while reducing energy consumption by approximately 50.05\% by eliminating continuous pneumatic actuation during grasp maintenance.
Overall, this work demonstrates a practical and energy-efficient approach to rigid–compliant integrated gripper design for agricultural harvesting. Future work will focus on improving the durability of the soft actuators under long-cycle operation and incorporating sensory feedback for closed-loop force regulation to further enhance system robustness and autonomy.


\addtolength{\textheight}{-12cm}   





\vspace{12pt}
\color{red}

\end{document}